\documentclass[10pt,twocolumn,letterpaper]{article}

\usepackage[pagenumbers]{cvpr} 

\definecolor{cvprblue}{rgb}{0.21,0.49,0.74}
\usepackage[pagebackref,breaklinks,colorlinks,allcolors=cvprblue]{hyperref}
\usepackage{tabularx}
\usepackage{booktabs}
\usepackage{makecell}
\usepackage{comment}
\usepackage{multirow}

\def\paperID{*****} 
\def\confName{CVPR}
\def\confYear{2025}

\title{ZERO: Industry-ready Vision Foundation Model with Multi-modal Prompts}

\author{
Sangbum Choi\thanks{Equal contribution} \ \thanks{Corresponding author} \\
Superb AI\\
Seoul, South Korea\\
{\tt\small sbchoi@superb-ai.com}
\and
Kyeongryeol Go\footnotemark[1] \\
Superb AI\\
Seoul, South Korea\\
{\tt\small krgo@superb-ai.com}
\and
Taewoong Jang\footnotemark[1] \\
Superb AI\\
Seoul, South Korea\\
{\tt\small twjang@superb-ai.com}
}

\begin{document}
\maketitle
\begin{abstract}
Foundation models have revolutionized AI, yet they struggle with zero-shot deployment in real-world industrial settings due to a lack of high-quality, domain-specific datasets. To bridge this gap, Superb AI introduces ZERO, an industry-ready vision foundation model that leverages multi-modal prompting (textual and visual) for generalization without retraining. Trained on a compact yet representative 0.9 million annotated samples from a proprietary billion-scale industrial dataset, ZERO demonstrates competitive performance on academic benchmarks like LVIS-Val and significantly outperforms existing models across 37 diverse industrial datasets. Furthermore, ZERO achieved 2nd place in the CVPR 2025 Object Instance Detection Challenge and 4th place in the Foundational Few-shot Object Detection Challenge, highlighting its practical deployability and generalizability with minimal adaptation and limited data. To the best of our knowledge, ZERO is the first vision foundation model explicitly built for domain-specific, zero-shot industrial applications.
\end{abstract}    
\section{Introduction}
\label{sec:intro}

Historically, improving model performance required deep domain expertise to inject task-specific inductive biases into model architectures. However, the advent of large-scale datasets has shifted this paradigm. Empirical evidence increasingly supports that stacking generic Transformer layers, rather than hand-crafting domain-specific modules, can yield superior results \cite{dosovitskiy2020image}. This shift has simplified model design and reduced reliance on manual heuristics.

This transition has given rise to foundation models, which have substantially accelerated progress across a wide range of AI tasks. Text-centric models like BERT \cite{devlin2019bert}, vision-centric models such as DINOv2 \cite{oquab2023dinov2}, and multi-modal models including CLIP \cite{radford2021learning} have demonstrated remarkable generalization capabilities. These advances have been fueled by the release of massive public datasets (e.g., SA-1B \cite{kirillov2023segment}, LAION-5B \cite{schuhmann2022laion}) and parameter-efficient fine-tuning methods like LoRA \cite{hu2022lora}, which enable fast adaptation with minimal computational cost.

Despite these advancements, current foundation models remain ill-suited for zero-shot deployment in real-world, domain-specific industrial settings. The datasets used to train them largely focus on general-purpose content, while high-quality, domain-specific datasets, especially in areas such as manufacturing, logistics, and medical imaging, are scarce due to high annotation costs and long-tail data distributions. Generative models can support data augmentation, but their benefits are often constrained by diminishing returns, as predicted by scaling laws \cite{yamaguchi2023limitation, fan2024scaling}. Recent work on active learning, data valuation, and subset selection \cite{go2023transferable, vo2024automatic} aims to mitigate these issues, but the challenge of curating effective domain-specific datasets persists as a core bottleneck to industrial AI adoption.

To address this gap, we present ZERO, a vision foundation model designed to be both industry-ready and data-efficient. Unlike traditional foundation models that rely heavily on large, general-purpose pretraining and require substantial downstream adaptation, ZERO leverages multi-modal prompting—both textual and visual—to enable generalization across tasks and domains without retraining. This approach allows ZERO to be directly deployed in production environments, significantly reducing the time and cost associated with AI integration in enterprise settings.

Developed by Superb AI, ZERO builds on our extensive experience in supporting end-to-end AI development pipelines for enterprise customers across diverse verticals. Leveraging a proprietary, billion-scale industrial dataset spanning manufacturing, retail, logistics, security, and surveillance, we curated a compact yet highly representative dataset of 0.9 million annotated samples. Despite its relatively modest size compared to other vision foundation training corpora, this dataset captures the complexity and variability inherent in real-world industrial scenarios.

Our experiments demonstrate that ZERO achieves competitive performance on LVIS-Val, the widely used academic benchmark, while outperforming existing models across 37 multi-domain industrial datasets. Moreover, ZERO achieved top ranks in competitive open-world settings, placing 2nd in the Object Instance Detection Challenge and 4th (honorable mention) in the Foundational Few-shot Object Detection (FSOD) Challenge, both held at the CVPR 2025 open world vision workshop. These results underscore the unique position of ZERO as a foundation model that is not only powerful and generalizable but also practically deployable with minimal adaptation and limited data. To our knowledge, ZERO is the first vision foundation model explicitly built for domain-specific, zero-shot industrial applications, bridging the persistent gap between academic innovation and real-world enterprise needs.
\section{Related Work}
\label{sec:relatedwork}

\subsection{Open-vocabulary object detection}
Recent advancements in contrastive learning have enabled models to relate images and text by mapping text and image embeddings into a shared semantic space \cite{radford2021learning,zhai2023sigmoid,tschannen2025siglip}.  This breakthrough has allowed open-vocabulary object detectors to be trained successfully on large-scale web-based image datasets without extensive relabeling. 

Pioneering works, including OWL-ViT \cite{minderer2022simple} and GLIP \cite{li2022grounded}, used contrastive learning methods to align image and text embeddings. GroundingDINO \cite{liu2024grounding} significantly improved performance by integrating modern object detection architectures with language models in a dual-encoder framework.  GroundingDINO became the foundation for subsequent research, including MM-GroundingDINO \cite{zhao2024open}, which extended capabilities to multi-modal inputs, and Grounded-SAM \cite{ren2024grounded}, which combined SAM \cite{ravi2024sam} for segmentation. Most recently, LLM-Det \cite{fu2025llmdet} has advanced this research direction by incorporating large language models, enabling models to understand more sophisticated natural language queries.

\subsection{Object detection with visual prompts}

Another line of research extends open-vocabulary object detection by incorporating visual prompts to alleviate the inherent ambiguity in text prompts. This approach includes DINOv \cite{li2024visual}, T-Rex \cite{jiang2023trex}, and T-Rex2 \cite{jiang2024t}. DINOv, built on top of DINO \cite{zhang2022dino}, enables visual in-context prompting, where visual prompts come from the same input images in which the model searches for objects. T-Rex and T-Rex2 relax this constraint of in-context prompting, allowing models to use visual examples from any image source. DINO-X\cite{ren2024dino} further pushed the boundaries by enabling models to present object masks and keypoints. It also introduced universal prompting, where the model locates all objects without user prompts.

ZERO was initially built on an open-vocabulary object detection model and was progressively extended to support not only text prompts but also visual prompts. In Section~\ref{sec:zero_train}, we present a training strategy that enables the model to effectively leverage visual prompts while preserving and gradually enhancing its pretrained ability to interpret text prompts.

\subsection{Real-time visual grounding}

One problem with previous models, despite being powerful, is that they require substantial memory and computational resources, which hinders practical deployment and real-time inference in industry. To address this problem, several works have implemented open-set object detectors based on YOLO architectures. YOLO-World \cite{cheng2024yolo} was first proposed for this purpose, and YOLOE \cite{wang2025yoloerealtimeseeing} further improved performance in terms of mAP through more sophisticated architectural design. Although ZERO is not intended for real-time inference, we discuss our efficiency-oriented design choices for deployment in Section~\ref{sec:zero_train_inf}.

\subsection{Referring expression comprehension}
Recent improvements in vision-language models (VLMs) have opened up a new research direction involving more complex logical reasoning for object detection. Compared to previous works, this line of research focuses more on filtering detected objects using VLMs' capabilities, enabling models to account for positional relationships between objects and object attributes. These works constructed their own datasets to train their models by integrating outputs from existing models and manual labeling into a loop. ChatRex \cite{jiang2024chatrextamingmultimodalllm} was the first attempt to merge VLMs with object detectors, and Rex-Seek \cite{jiang2025referringperson} focuses more on processing human attribute queries. Rex-Thinker \cite{jiang2025rexthinkergroundedobjectreferring} further improved Rex-Seek by adding reasoning capabilities to the model.

As described in Section~\ref{sec:zero_data_lab}, ZERO leverages a vision-language model (VLM) solely for constructing an auto-labeling pipeline to obtain high-quality, richly annotated data; it is not used during model training or inference. Extending the use of VLMs to incorporate visual context for enhanced scene understanding and video analysis is left as future work.
\section{ZERO}
\label{sec:zero}

\begin{figure}[t]
    \centering
    \includegraphics[width=\linewidth]{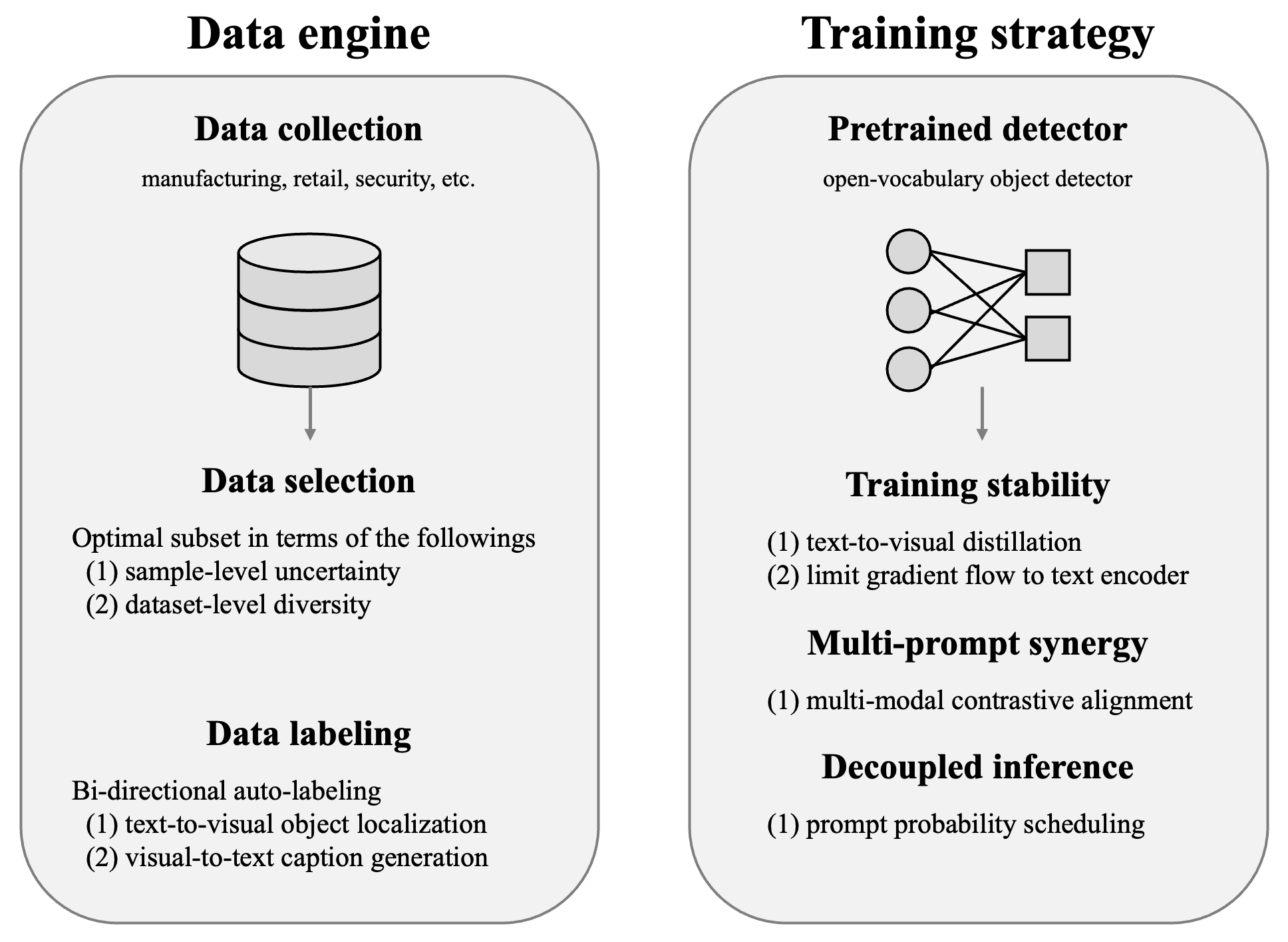}
    \caption{
    Overview of the proposed ZERO training pipeline. The pipeline is composed of a data engine (left), which constructs a compact and richly annotated industrial dataset through collection, selection, and pseudo-labeling; and a training strategy (right) that progressively adapts a pretrained open-vocabulary detector using distillation and alignment, ultimately enabling decoupled inference with either text or visual prompt.
    }
    \label{fig:pipeline}
\end{figure}

To enable efficient and robust training of ZERO, we design a comprehensive methodology centered around two key pillars: a data engine and a training strategy. The data engine systematically constructs a compact yet richly annotated dataset through advanced collection, selection, and bidirectional pseudo-labeling pipelines. Complementing this, our training strategy progressively adapts pretrained single-modal models to support multi-modal prompts by leveraging carefully orchestrated distillation and contrastive learning techniques. Together, these components form the foundation for a versatile vision foundation model that excels in both detection and referring tasks while maintaining inference efficiency. Please refer to the overall pipeline in Figure~\ref{fig:pipeline}.

\subsection{Data engine}
\label{sec:zero_data}

To effectively train ZERO, we designed a data engine that comprises three core components: collection, selection, and labeling. This engine is built with the goal of constructing a compact yet highly representative dataset tailored for data-efficient training of vision foundation models in real-world industrial settings.

\subsubsection{Data collection}

Superb AI has accumulated a large-scale repository of industrial domain data through years of supporting end-to-end AI development pipelines for enterprise customers. This proprietary data asset spans diverse verticals, including manufacturing, retail, logistics, security, and surveillance, and serves as the foundation for training ZERO.

\subsubsection{Data selection}

In building the training dataset, we employed advanced data selection techniques developed through years of operating commercial data curation services. The selection process is formulated as a multi-objective optimization task that jointly considers sample-level uncertainty and dataset-level diversity. To ensure that the resulting subset remains informative and transferable across a wide range of models and domains, we leveraged embedding representations from multiple foundation models. Empirical results demonstrate that our selection strategy can achieve near-equivalent training performance using only a fraction of the original dataset. Based on this, we curated a dataset of 0.9 million high-quality samples for ZERO training. While this scale is significantly smaller than that of other vision foundation models, it highlights the maturity of Superb AI’s data-centric approach and demonstrates that foundation models can be developed efficiently without requiring massive resources, thus offering a viable path even for startups and academic institutions.

\subsubsection{Data labeling}
\label{sec:zero_data_lab}

To maximize the utility of each training sample and enable broader task generalization, we developed a customized auto-labeling pipeline. This pipeline not only enhances per-sample learning efficiency but also expands the annotation space to include referring expression comprehension, which involves complex linguistic structures beyond traditional detection tasks. Unlike recent VLM-based approaches that improve referring performance at the expense of detection accuracy, our pipeline is designed to enrich the annotation quality in both coarse and fine-grained levels, thereby enabling a single model to handle both detection and referring tasks effectively.

The pipeline integrates two complementary directions. First, we extract captions from images using InternVL3 \cite{zhu2025internvl3} and identify noun phrases via SpaCy \cite{Honnibal_spaCy_Industrial-strength_Natural_2020}. These noun phrases are filtered using WordNet \cite{miller1995wordnet} to isolate expressions corresponding to physical objects. Using a preliminary version of ZERO trained on ground-truth annotations, we then aggregate inference results guided by both textual and visual prompts to localize bounding boxes for each phrase. Second, we reverse the process by generating object proposals with the automatic mask generation by SAM2 \cite{ravi2024sam} and extracting corresponding captions using Qwen2.5-VL \cite{Qwen2.5-VL}. In this step, the captioning focuses on visual attributes such as color, shape, and material, allowing for richer object-level descriptions. To maintain label quality, we filter out instances with extremely small bounding boxes and apply SigLIP2 \cite{tschannen2025siglip} to assess alignment between cropped image regions and their generated captions. Low-alignment pairs are removed from the training set.

Altogether, this data engine integrates and extends strategies commonly used in the construction of foundation model datasets. Our empirical analysis shows that the bidirectional labeling process effectively yields high-quality supervision signals across different granularities and plays a critical role in enabling ZERO to generalize across a wide range of industrial domains.

\subsection{Training strategy}
\label{sec:zero_train}

While numerous open-source models for the visual grounding task exist, they are typically limited to supporting a single type of prompt, most often textual. Some models like YOLOE \cite{wang2025yoloe} support visual prompting, but still underperform compared to modern open-vocabulary detectors \cite{jiang2024t, ren2024dino}. To overcome this limitation without redesigning the base model architecture, we focus on extending pretrained single-modal models to support multiple prompt modalities. In summary, our approach distills the open-vocabulary detection capabilities of a pretrained model in the early stages and subsequently enhances performance through multi-prompt synergy. The final trained model is efficient at inference time, requiring only the contrastive prompt encoders.

\subsubsection{Progressive adaptation to multi-prompt}

Naively appending a visual prompt encoder can lead to performance degradation, particularly during early training stages, due to the representational mismatch with the pretrained text prompt encoder. To mitigate this, we begin training with text-only prompts and employ an InfoNCE loss \cite{oord2018representation} to align the visual prompt encoder with the pretrained text prompt encoder. To prevent degradation of the text embeddings, we limit gradient propagation so that distillation happens only in one direction: from the text encoder to the visual encoder.

Relying solely on unidirectional distillation for training stability can limit the synergy between text and visual prompts. To address this, we incorporate a contrastive backbone, such as CLIP \cite{radford2021learning}, which inherently supports text-visual alignment. In our framework, the visual prompt encoder is realized through the contrastive visual encoder, and a new contrastive text encoder is introduced to gradually replace the original pretrained text encoder.

This design yields three distinct prompt encoders: the \textit{pretrained text prompt encoder}, the \textit{contrastive visual prompt encoder}, and the \textit{contrastive text prompt encoder}. During training, we apply unidirectional distillation from the pretrained text encoder to the contrastive visual encoder to ensure stability. In parallel, we promote multi-modal synergy by enforcing bidirectional alignment between the contrastive visual and text encoders.

To gradually reduce reliance on the pretrained encoder, we adopt a probabilistic prompt scheduling strategy. During the first half of training, the usage probability of the pretrained text encoder decreases linearly from 1.0 to 0.0, while the probabilities for both contrastive prompt encoders increase from 0.0 to 0.5. These values remain fixed for the remainder of the training.

\subsubsection{Decoupled inference}
\label{sec:zero_train_inf}

An important note is that the purpose of the distillation and alignment losses is to promote a well-structured embedding space through multi-prompt interaction during training. During inference, however, it is not necessary to provide multiple prompts simultaneously, since the forward pass does not involve prompt-level interaction. The model can accept either a text prompt, a visual prompt, or both, with optional merging in an ensemble-like manner. Consequently, the pretrained text prompt encoder can be safely removed at deployment, and inference can proceed using only the contrastive backbone.

\section{Experiment}
\label{sec:experiment}

This section presents a comprehensive evaluation of the proposed ZERO model across a wide range of settings. Our experiments are organized along two main axes: (1) zero-shot performance on public benchmarks and industrial multi-domain datasets, and (2) results from competitive challenges at the CVPR 2025 Open World Vision workshop.

\begin{table*}[t]
\centering
\caption{Zero-shot performance on LVIS-Val benchmark and Multi-domain datasets.}
\label{tab:metric}
\begin{tabular}{lcccccc}
\toprule
\multirow{2}{*}{Model} & \multirow{2}{*}{Training dataset size} & \multicolumn{1}{c}{\textbf{LVIS-Val}} & \multicolumn{3}{c}{\textbf{Multi-domain dataset}} \\
\cmidrule(lr){3-3} \cmidrule(lr){4-6}
& & Visual-I AP & Text-G AP & Visual-G AP & Max AP \\
\midrule
T-Rex2 \cite{jiang2024t}        & 6.5M            & \textbf{62.5} & -     & 13.0   & -     \\
Florence-2 \cite{xiao2024florence}    & 126M            & -    & 4.3   & -      & -     \\
YOLOE \cite{wang2025yoloe}         & 1.4M            & 50.5 & 7.5   & 12.0   & 15.0  \\
OV-DINO \cite{wang2024ov}       & 2.4M            & -    & 10.2  & -      & -     \\
DINO-X \cite{ren2024dino}        & 20M $\sim$ 100M & -    & 12.8  & -      & -     \\
OWLv2 \cite{minderer2023scaling}         & 2.1M            & -    & 12.9  & -      & -     \\
\midrule
\textbf{ZERO}  & 0.9M  & 60.1 & \textbf{24.5} & \textbf{18.5} & \textbf{29.1} \\
\bottomrule
\end{tabular}
\end{table*}

\subsection{Zero-shot performance}

We evaluate zero-shot detection performance under three prompting protocols introduced in \citet{jiang2024t}. Text-G corresponds to the standard open-vocabulary setting, where object categories are specified using only text prompts (i.e., class names) without any visual exemplars. Visual-G employs category-level visual prompts derived by averaging the embeddings of $N$ training exemplars per class, simulating scenarios where a few reference images are available. Visual-I represents an interactive setting in which a single instance crop from the test image serves as the visual prompt to detect similar objects within the same image. While this assumes access to ground-truth regions at test time, it reflects practical applications such as interactive annotation. Results are summarized in Table~\ref{tab:metric}. Note that cells marked with ‘–’ indicate that the corresponding prompt type is not supported for the given model.

\subsubsection{Academic benchmark}

ZERO demonstrates strong performance on the widely used LVIS-Val benchmark, achieving a Visual-I AP score of 60.1. This outperforms comparable open-source models such as YOLOE \cite{wang2025yoloe} (50.5) and approaches the state-of-the-art T-Rex2 \cite{jiang2024t} (62.5). Remarkably, ZERO attains this level of accuracy using only 0.9M training examples—significantly fewer than other models that rely on tens or even hundreds of millions of samples—highlighting its data efficiency and practical scalability.

\subsubsection{Industrial benchmark}

To evaluate generalization in real-world settings, we assess ZERO on 37 diverse industrial datasets spanning domains such as healthcare, autonomous driving, retail, and gaming. In this multi-domain dataset, we report three aggregate metrics. Text-G and Visual-G scores are first computed per domain and then averaged across domains. The Max AP metric is obtained by selecting the higher value between the Text-G and Visual-G scores for each domain, followed by averaging these maximum values across all domains. Together, these metrics provide a comprehensive view of the model’s performance across a wide range of scenarios. Across all domains, ZERO consistently outperforms previous models by a substantial margin.

\begin{figure*}[t]
    \centering
    \begin{subfigure}{\textwidth}
        \centering
        \includegraphics[width=0.8\linewidth]{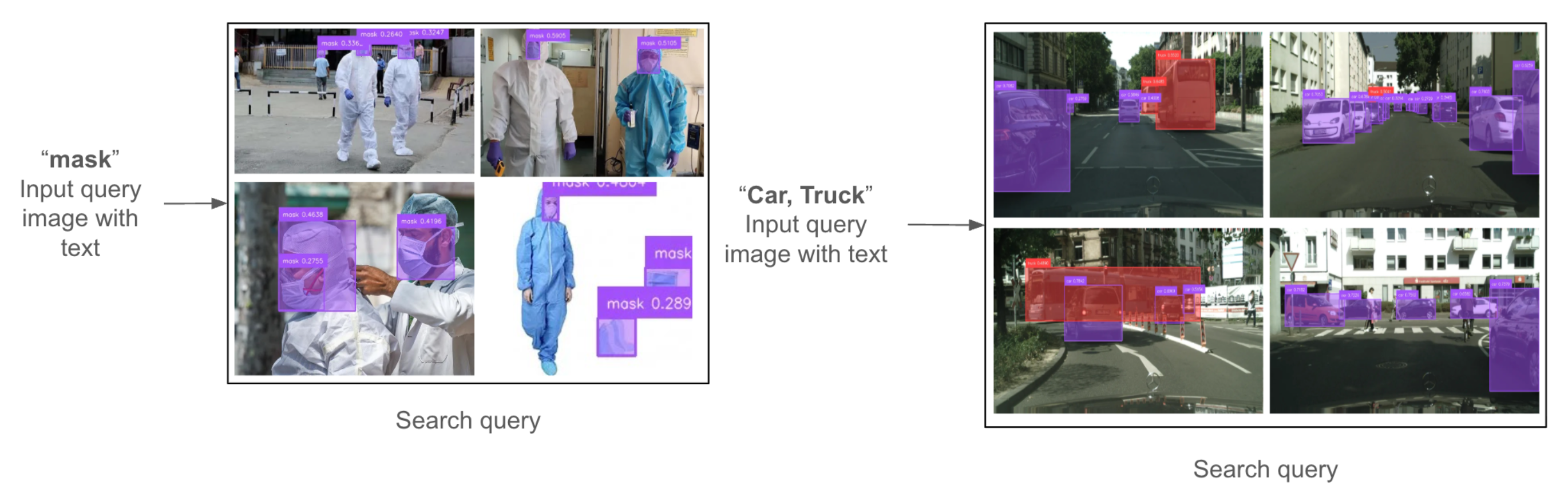}
        \caption{Text-prompt-based zero-shot detection examples}
        \label{fig:text_prompt}
    \end{subfigure}
    
    \begin{subfigure}{\textwidth}
        \centering
        \includegraphics[width=0.8\linewidth]{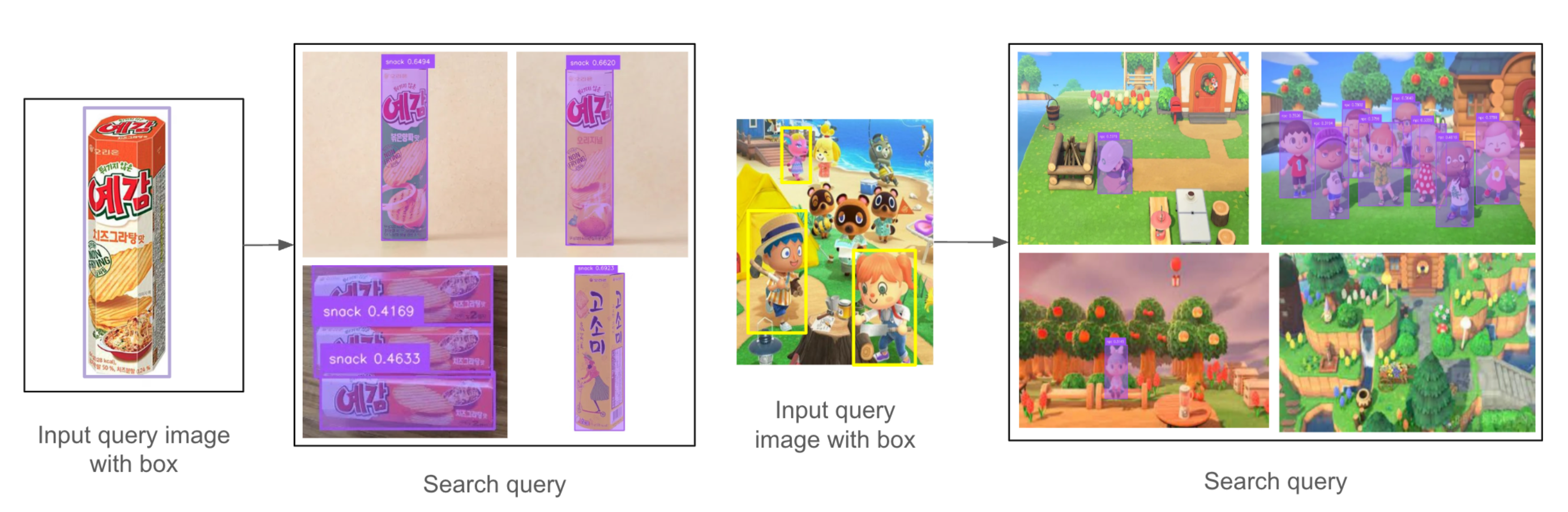}
        \caption{Visual-prompt-based zero-shot detection examples}
        \label{fig:visual_prompt}
    \end{subfigure}

    \caption{Qualitative examples of ZERO in zero-shot detection settings using different types of prompts.}
    \label{fig:zero_shot_examples}
\end{figure*}

\subsubsection{Qualitative evaluation}

Beyond strong quantitative results, ZERO also demonstrates impressive qualitative performance across diverse visual domains. For example, it precisely segments masks of varying shapes and sizes in medical images, reliably detects cars and trucks in complex traffic scenes for autonomous driving, and maintains high accuracy in retail and gaming environments despite visual clutter and style variations. Representative visual examples are presented in Figure~\ref{fig:zero_shot_examples}.

\subsection{CVPR 2025 challenge}

To evaluate the real-world utility of ZERO as a general-purpose vision foundation model, we participated in two practically motivated object detection challenges: the Object Instance Detection (InsDet) Challenge and the Foundational Few-shot Object Detection (FSOD) Challenge. These benchmarks assess complementary aspects of visual recognition—instance-level retrieval in open-world scenarios and few-shot generalization across diverse industrial domains. Both tasks minimize reliance on predefined categories or extensive supervision, aligning well with ZERO’s multi-modal, prompt-driven architecture. In this section, we describe our methods and report the results achieved by ZERO in each challenge.

\subsubsection{Object Instance Detection challenge}

InsDet challenge presents a practically motivated benchmark built on the recently introduced InsDet dataset. Unlike conventional object detection tasks that aim to recognize all instances of predefined categories, the task focuses on detecting specific object instances provided as visual exemplars captured from multiple viewpoints. This formulation is especially relevant for robotics and retail automation applications, such as assistive robots retrieving household items or fulfillment robots navigating cluttered shelves. The challenge thus offers a realistic and high-stakes testbed for evaluating models in open-world, instance-level detection scenarios.

ZERO, with its multi-modal prompt architecture and zero-shot adaptability, is well aligned with the nature of the task. Instead of relying on fixed category labels, ZERO can leverage visual prompts to identify target objects, making it particularly suitable for instance-level retrieval in unstructured environments. To this end, we started with a base pipeline using the inference outputs of ZERO.

From this baseline, we introduced a series of targeted improvements that reflect practical deployment constraints while maximizing performance under zero-shot conditions. First, category refinement was applied using a visual embedding model to align instance representations, combined with tiled inference to improve localization granularity. Next, to reduce spurious background predictions, a common issue in open-world detection, we trained a lightweight background classifier. Batched Non-Maximum Suppression (NMS) was then employed to handle overlapping predictions at scale efficiently. Finally, bounding box quality was enhanced using SAM2-based refinement \cite{ravi2024sam}. The progressive enhancements are reported in Table~\ref{tab:instdet} and demonstrate the effectiveness of modular inference-time improvements of ZERO that do not require additional training or fine-tuning.

\begin{table}[t]
\centering
\begin{tabularx}{\columnwidth}{X c c}
\toprule
Method & AP & Gain \\
\midrule
Standard inference with ZERO & 48.5 & -- \\
+ Category refinement & 52.8 & +4.3 \\
+ Background filtering & 58.0 & +5.2 \\
+ Batched NMS & 69.1 & +11.1 \\
+ Box refinement with SAM2 \cite{ravi2024sam} & \textbf{70.3} & +1.2 \\
\bottomrule
\end{tabularx}
\caption{Progressive improvements of ZERO on the InsDet Challenge. Each row adds a new component to the previous setting.}
\label{tab:instdet}
\end{table}

A key innovation in our participation was the data preparation strategy. While InsDet supports synthetic scene generation by pasting instance crops into varied backgrounds, we extended this approach using generative AI services to synthesize visually rich and diverse scenes. By applying ZERO to these generated images, we could identify background-class objects that were not originally present in the InsDet dataset. These new samples were then used to train the background classifier, effectively filtering irrelevant detections and enriching the dataset with semantically novel content. This pipeline significantly improved instance discrimination and reduced false positives, demonstrating a practical synergy between generative models and vision foundation models for open-world instance detection.

Overall, our results, placing 2nd in the challenge, highlight ZERO’s strength as a robust, generalizable, and industry-ready foundation model. The modular improvements we propose are inference-time only, lightweight, and easily extensible, illustrating that ZERO can be rapidly adapted to domain-specific requirements without retraining. These attributes make ZERO especially suitable for real-world applications where flexibility, efficiency, and zero-shot generalization are essential.

\subsubsection{Foundation Few-shot Object Detection challenge}

Foundational FSOD challenge \citep{madan2024revisiting, robicheaux2025roboflow100} is designed to rigorously evaluate models’ capabilities in performing object detection with extremely limited annotations across diverse and heterogeneous industrial domains. Using the RF20VL-fsod dataset, which encompasses 20 distinct domains, participants are tasked with detecting objects given only 10 bounding box annotations per category. This sparse labeling setup realistically simulates practical constraints in industrial settings where exhaustive annotation is costly or infeasible.

Each category is supplemented with detailed noun phrase descriptions to clarify ambiguous or domain-specific terminology, supporting better semantic understanding. The challenge allows pretraining on external datasets but mandates that fine-tuning be conducted solely on the RF20VL-fsod dataset, emphasizing adaptation to the target domains under few-shot conditions. The final performance is measured by the mean average precision (mAP) at IoU thresholds ranging from 0.5 to 0.95, averaged across all domains, thereby assessing both detection accuracy and generalization.

Unlike zero-shot benchmarks, the Foundational FSOD challenge explicitly tests a model’s ability to adapt efficiently and robustly to new domains with minimal supervision, making it a critical assessment for foundation models intended for industrial deployment. To address these requirements, we designed a training pipeline that maximizes prompt diversity at both textual and visual levels.

We introduce extensive prompt diversity to mitigate overfitting. At the text level, we employ LLaMA-3-8B-Instruct to paraphrase category descriptions into concise noun phrases. The paraphrasing process is guided by a carefully crafted prompt template in Table~\ref{tab:paraphrase_prompt} that instructs the model to preserve semantic integrity, avoid redundancy, and generate unambiguous and distinct definitions. This augmentation not only increases the lexical variety of prompts but also improves the model’s ability to distinguish fine-grained categories. In addition to positive prompts, we include negative textual prompts to encourage a better discriminative embedding space obtained by contrastive learning. Note that the number of negative prompts is set with caution to avoid introducing label noise.

\begin{table}[t]
\centering
\begin{tabularx}{\columnwidth}{X}
\toprule
\textbf{Instruction} \\
\midrule
You are an assistant specialized in generating concise noun-phrase definitions by paraphrasing. You will be given a list of terms in the format \texttt{[term] = [definition]}. For each term, return a corresponding line in the format \texttt{[term] = [paraphrased definition]}. \\
Your paraphrased definitions must: \\
\hspace{1em}1. Be concise and written as noun phrases. \\
\hspace{1em}2. Preserve the original meaning and context. \\
\hspace{1em}3. Clearly distinguish each term from the others. \\
\hspace{1em}4. Follow the same line-by-line format as the input. \\
Do not add or omit any terms. \\
\bottomrule
\end{tabularx}
\caption{Instruction used for text prompt augmentation.}
\label{tab:paraphrase_prompt}
\end{table}

At the visual level, we apply both in-image and out-image prompting strategies. In-image visual prompts use objects co-occurring in the same image, reflecting a convention common in large-scale training in the vision foundation model and aiding contextual reasoning. Out-image visual prompts, by contrast, introduce objects of the same category from other images, fostering generalization and reducing reliance on narrow contextual cues.

Alongside prompt engineering, we implement a conservative pseudo-labeling strategy: only model predictions with high confidence scores are added as pseudo-labels for instances of unlabeled categories in the train split. This strict filtering ensures the integrity of additional supervision while preventing error accumulation from low-quality labels.

During inference, we adopt several techniques to improve detection performance and stability. Test-time augmentation (TTA) is applied by default using a combination of image resizing at multiple scales (0.8×, 1.0×, 1.2×) and horizontal flipping, effectively exposing the model to various spatial configurations. We also perform a category-wise threshold search, tuning the confidence threshold for each class individually to optimize the mean average precision (mAP) in the validation split. To further enhance robustness, we considered the ensemble of predictions from both text and visual prompts.

Due to the distinctiveness of each domain, no universal strategy consistently outperforms others. Therefore, we obtained various model checkpoints by turning on and off the proposals for training and inference, and selected the best checkpoint guided by the performance on a validation split. For each dataset from the 20 distinct domains, the best checkpoint is selected based on combinations of four factors in Table~\ref{tab:factor_selection}. It is important to note that, similar to the training split, the validation split is only partially annotated. Therefore, during checkpoint evaluation, we restrict predictions to the categories known to be present in each image. In contrast, for test split inference intended for final submission, predictions are obtained across all categories.

\begin{table}[t]
\centering
\begin{tabularx}{\columnwidth}{lX}
\toprule
\textbf{Factor} & \textbf{Options} \\
\midrule
Text prompt &
\makecell[l]{
\texttt{original} \\
\texttt{original + augmented}
} \\
\midrule
Visual prompt &
\makecell[l]{
\texttt{in-image} \\
\texttt{out-image}
} \\
\midrule
Annotations &
\makecell[l]{
\texttt{original} \\
\texttt{original + pseudo-labeled}
} \\
\midrule
Inference &
\makecell[l]{
\texttt{text} \\
\texttt{visual} \\
\texttt{text + visual}
} \\
\bottomrule
\end{tabularx}
\caption{Factors considered for checkpoint selection.}
\label{tab:factor_selection}
\end{table}

Consequently, we achieved 4th place and received an honorable mention from the challenge organizers, highlighting ZERO’s strong potential as an industry-ready vision foundation model.
\section{Conclusion}
\label{sec:conclusion}

This paper introduces ZERO, a multi-modal prompt object detection model specifically designed for industry-specific zero-shot deployment. ZERO leverages both textual and visual prompts to generalize across various tasks and domains without requiring retraining. Developed by Superb AI, ZERO utilizes a proprietary industrial dataset of one billion images to demonstrate the potential of prompt-driven, data-centric AI for scalable and adaptive object detection in industrial environments. ZERO achieved strong performance in the CVPR 2025 Instance Object Detection and Foundational Few-Shot Object Detection Challenges, placing 2nd and 4th, respectively. ZERO offers practical advantages in terms of flexibility, efficiency, and real-world applicability, and is the first vision foundation model explicitly built for domain-specific, zero-shot industrial applications, bridging the persistent gap between academic innovation and real-world enterprise needs.
{
    \small
    \bibliographystyle{ieeenat_fullname}
    \bibliography{main}
}


\end{document}